\documentclass[a4paper,conference]{IEEEtran}
\title{Tarsier: Evolving Noise Injection in Super-Resolution GANs}
\usepackage[utf8]{inputenc}
\usepackage[T1]{fontenc}
\usepackage{booktabs} 
\usepackage{todonotes}
\usepackage{enumitem}
\usepackage{amsmath}

\usepackage{graphbox} 
\usepackage{adjustbox}

\newcommand{\omitme}[1]{}
\usepackage[caption = false]{subfig}
\usepackage{tikz}
\usepackage{pgfplots}
\usepackage{float}
\usetikzlibrary{spy,calc}
\usepackage{hyperref}
\def\MGAN{\textsc{nESRGAN+}}
\def\dcma{Diagonal CMA}
\def\ESRGAN{\textsc{ESRGAN}}

\def\K512{K}
\def\koncept{Koncept512}
\def\EnhanceNet{\textsc{EnhanceNet}}
\def\tarsier{Tarsier}

\def\discr{D}
 
 \newcommand{\mtb}[1]{\textcolor{black}{#1}}
 \newcommand{\otc}[1]{\textcolor{black}{#1}}
  \newcommand{\br}[1]{\textcolor{black}{#1}}

 \newcommand{\otcicpr}[1]{\textcolor{black}{#1}}
  \newcommand{\bricpr}[1]{\textcolor{black}{#1}}
\newcommand{\ccci}[1]{\textcolor{black}{#1}}

\newcommand{\ccc}[1]{\textcolor{black}{#1}}
\newcommand{\brthree}[1]{\textcolor{black}{#1}}

\newcommand{\brfour}[1]{\textcolor{black}{#1}}
 \newcommand{\otcyo}[1]{\textcolor{black}{#1}}
\newcommand{\cc}[1]{\textcolor{black}{#1}}

\newcommand{\brtwo}[1]{\textcolor{black}{#1}}

\author{\IEEEauthorblockN{Baptiste Roziere\IEEEauthorrefmark{1}\IEEEauthorrefmark{2},
Nathanaël Carraz Rakotonirina\IEEEauthorrefmark{3},
Vlad Hosu\IEEEauthorrefmark{4},
Andry Rasoanaivo\IEEEauthorrefmark{3},
Hanhe Lin\IEEEauthorrefmark{4}}
\ccci{Camille Couprie}\IEEEauthorrefmark{1} and
Olivier Teytaud\IEEEauthorrefmark{1}
\IEEEauthorblockA{\IEEEauthorrefmark{1}Facebook AI Research}
\IEEEauthorblockA{\IEEEauthorrefmark{2}Paris-Dauphine University}
\IEEEauthorblockA{\IEEEauthorrefmark{2}Laboratoire d’Informatique et Mathématiques, Université d’Antananarivo}
\IEEEauthorblockA{\IEEEauthorrefmark{3}University of Konstanz}
}
\newif\ifblackandwhitecycle
\gdef\patternnumber{0}

\pgfkeys{/tikz/.cd,
    zoombox paths/.style={
        draw=orange,
        very thick
    },
    black and white/.is choice,
    black and white/.default=static,
    black and white/static/.style={
        draw=white,
        zoombox paths/.append style={
            draw=white,
            postaction={
                draw=black,
                loosely dashed
            }
        }
    },
    black and white/static/.code={
        \gdef\patternnumber{1}
    },
    black and white/cycle/.code={
        \blackandwhitecycletrue
        \gdef\patternnumber{1}
    },
    black and white pattern/.is choice,
    black and white pattern/0/.style={},
    black and white pattern/1/.style={
            draw=white,
            postaction={
                draw=black,
                dash pattern=on 2pt off 2pt
            }
    },
    black and white pattern/2/.style={
            draw=white,
            postaction={
                draw=black,
                dash pattern=on 4pt off 4pt
            }
    },
    black and white pattern/3/.style={
            draw=white,
            postaction={
                draw=black,
                dash pattern=on 4pt off 4pt on 1pt off 4pt
            }
    },
    black and white pattern/4/.style={
            draw=white,
            postaction={
                draw=black,
                dash pattern=on 4pt off 2pt on 2 pt off 2pt on 2 pt off 2pt
            }
    },
    zoomboxarray inner gap/.initial=5pt,
    zoomboxarray columns/.initial=2,
    zoomboxarray rows/.initial=2,
    subfigurename/.initial={},
    figurename/.initial={zoombox},
    zoomboxarray/.style={
        execute at begin picture={
            \begin{scope}[
                spy using outlines={%
                    zoombox paths,
                    width=\imagewidth / \pgfkeysvalueof{/tikz/zoomboxarray columns} - (\pgfkeysvalueof{/tikz/zoomboxarray columns} - 1) / \pgfkeysvalueof{/tikz/zoomboxarray columns} * \pgfkeysvalueof{/tikz/zoomboxarray inner gap} -\pgflinewidth,
                    height=\imageheight / \pgfkeysvalueof{/tikz/zoomboxarray rows} - (\pgfkeysvalueof{/tikz/zoomboxarray rows} - 1) / \pgfkeysvalueof{/tikz/zoomboxarray rows} * \pgfkeysvalueof{/tikz/zoomboxarray inner gap}-\pgflinewidth,
                    magnification=3,
                    every spy on node/.style={
                        zoombox paths
                    },
                    every spy in node/.style={
                        zoombox paths
                    }
                }
            ]
        },
        execute at end picture={
            \end{scope}
            \node at (image.north) [anchor=north,inner sep=0pt] {\phantomimage};
            \node at (zoomboxes container.north) [anchor=north,inner sep=0pt] {\phantomimage};
     \gdef\patternnumber{0}
        },
        spymargin/.initial=0.5em,
        zoomboxes xshift/.initial=1,
        zoomboxes right/.code=\pgfkeys{/tikz/zoomboxes xshift=1},
        zoomboxes left/.code=\pgfkeys{/tikz/zoomboxes xshift=-1},
        zoomboxes yshift/.initial=0,
        zoomboxes above/.code={
            \pgfkeys{/tikz/zoomboxes yshift=1},
            \pgfkeys{/tikz/zoomboxes xshift=0}
        },
        zoomboxes below/.code={
            \pgfkeys{/tikz/zoomboxes yshift=-1},
            \pgfkeys{/tikz/zoomboxes xshift=0}
        },
        caption margin/.initial=4ex,
    },
    adjust caption spacing/.code={},
    image container/.style={
        inner sep=0pt,
        at=(image.north),
        anchor=north,
        adjust caption spacing
    },
    zoomboxes container/.style={
        inner sep=0pt,
        at=(image.north),
        anchor=north,
        name=zoomboxes container,
        xshift=\pgfkeysvalueof{/tikz/zoomboxes xshift}*(\imagewidth+\pgfkeysvalueof{/tikz/spymargin}),
        yshift=\pgfkeysvalueof{/tikz/zoomboxes yshift}*(\imageheight+\pgfkeysvalueof{/tikz/spymargin}+\pgfkeysvalueof{/tikz/caption margin}),
        adjust caption spacing
    },
    calculate dimensions/.code={
        \pgfpointdiff{\pgfpointanchor{image}{south west} }{\pgfpointanchor{image}{north east} }
        \pgfgetlastxy{\imagewidth}{\imageheight}
        \global\let\imagewidth=\imagewidth
        \global\let\imageheight=\imageheight
        \gdef\columncount{1}
        \gdef\rowcount{1}
        
    },
    image node/.style={
        inner sep=0pt,
        name=image,
        anchor=south west,
        append after command={
            [calculate dimensions]
            node [image container,subfigurename=\pgfkeysvalueof{/tikz/figurename}-image] {\phantomimage}
            node [zoomboxes container,subfigurename=\pgfkeysvalueof{/tikz/figurename}-zoom] {\phantomimage}
        }
    },
    color code/.style={
        zoombox paths/.append style={draw=#1}
    },
    connect zoomboxes/.style={
    spy connection path={\draw[draw=none,zoombox paths] (tikzspyonnode) -- (tikzspyinnode);}
    },
    help grid code/.code={
        \begin{scope}[
                x={(image.south east)},
                y={(image.north west)},
                font=\footnotesize,
                help lines,
                overlay
            ]
            \foreach \x in {0,1,...,9} {
                \draw(\x/10,0) -- (\x/10,1);
                \node [anchor=north] at (\x/10,0) {0.\x};
            }
            \foreach \y in {0,1,...,9} {
                \draw(0,\y/10) -- (1,\y/10);                        \node [anchor=east] at (0,\y/10) {0.\y};
            }
        \end{scope}
    },
    help grid/.style={
        append after command={
            [help grid code]
        }
    },
    }

\newcommand\phantomimage{%
    \phantom{%
        \rule{\imagewidth}{\imageheight}%
    }%
}
\newcommand\zoombox[2][]{
    \begin{scope}[zoombox paths]
        \pgfmathsetmacro\xpos{
            (\columncount-1)*(\imagewidth / \pgfkeysvalueof{/tikz/zoomboxarray columns} + \pgfkeysvalueof{/tikz/zoomboxarray inner gap} / \pgfkeysvalueof{/tikz/zoomboxarray columns} ) + \pgflinewidth
        }
        \pgfmathsetmacro\ypos{
            (\rowcount-1)*( \imageheight / \pgfkeysvalueof{/tikz/zoomboxarray rows} + \pgfkeysvalueof{/tikz/zoomboxarray inner gap} / \pgfkeysvalueof{/tikz/zoomboxarray rows} ) + 0.5*\pgflinewidth
        }
        \edef\dospy{\noexpand\spy [
            #1,
            zoombox paths/.append style={
                black and white pattern=\patternnumber
            },
            every spy on node/.append style={#1},
            x=\imagewidth,
            y=\imageheight
        ] on (#2) in node [anchor=north west] at ($(zoomboxes container.north west)+(\xpos pt,-\ypos pt)$);}
        \dospy
        \pgfmathtruncatemacro\pgfmathresult{ifthenelse(\columncount==\pgfkeysvalueof{/tikz/zoomboxarray columns},\rowcount+1,\rowcount)}
        \global\let\rowcount=\pgfmathresult
        \pgfmathtruncatemacro\pgfmathresult{ifthenelse(\columncount==\pgfkeysvalueof{/tikz/zoomboxarray columns},1,\columncount+1)}
        \global\let\columncount=\pgfmathresult
        \ifblackandwhitecycle
            \pgfmathtruncatemacro{\newpatternnumber}{\patternnumber+1}
            \global\edef\patternnumber{\newpatternnumber}
        \fi
    \end{scope}
}

\begin{document}

\maketitle
\begin{abstract}
Super-resolution aims at increasing the resolution and level of detail within an image.
The current state of the art in general single-image super-resolution is held by \MGAN{}, which injects a Gaussian noise after each residual layer at training time. 
In this paper, we harness evolutionary methods to improve \MGAN{} by optimizing the noise injection at inference time.
More precisely, \omitme{instead of injecting no noise,} we use \dcma{} to optimize the injected noise according to a novel criterion combining quality assessment and realism.
Our results are validated \br{by} the PIRM perceptual score \brfour{and a human study}. \br{Our method outperforms \MGAN{} on several standard super-resolution datasets.}
\otc{\br{More generally,} our approach can be used to \br{optimize} any method based on noise injection.}
\end{abstract}

\begin{figure}[t]
\begin{minipage}{0.45\textwidth}
\begin{center}
\begin{tabular}{ccc}
\setlength{\tabcolsep}{0.5pt}
\includegraphics[trim={27.5 67.5 77.5 40.},clip,width=.26\textwidth]{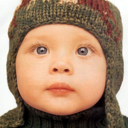}&
\includegraphics[trim={110 270 310 160},clip,width=.26\textwidth]{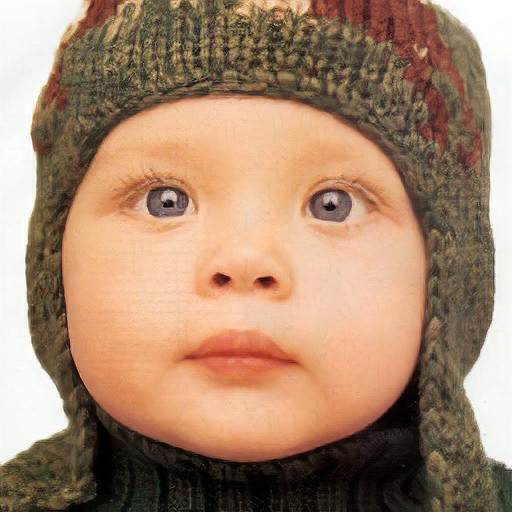}&
\includegraphics[trim={110 270 310 160},clip,width=.26\textwidth]{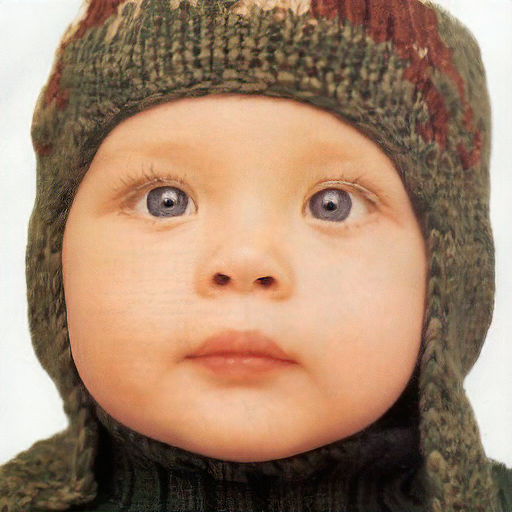}
\\
Low res.& {\small\EnhanceNet{}} & SRGAN \\
\includegraphics[trim={110 270 310 160},clip,width=.26\textwidth]{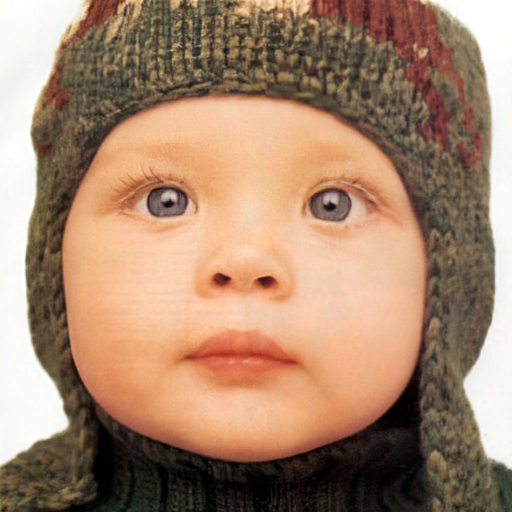}&
\includegraphics[trim={110 270 310 160},clip,width=.26\textwidth]{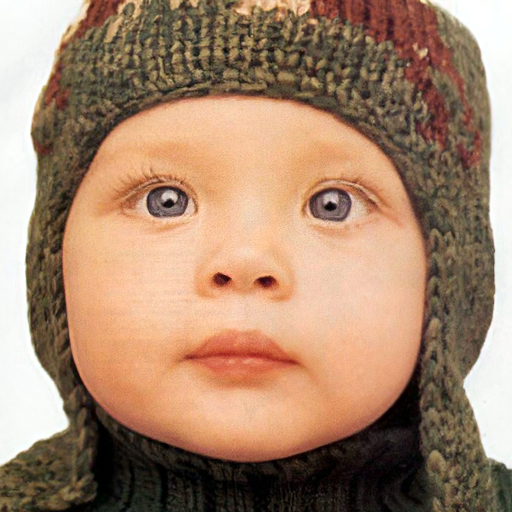}&
\includegraphics[trim={110 270 310 160},clip,width=.26\textwidth]{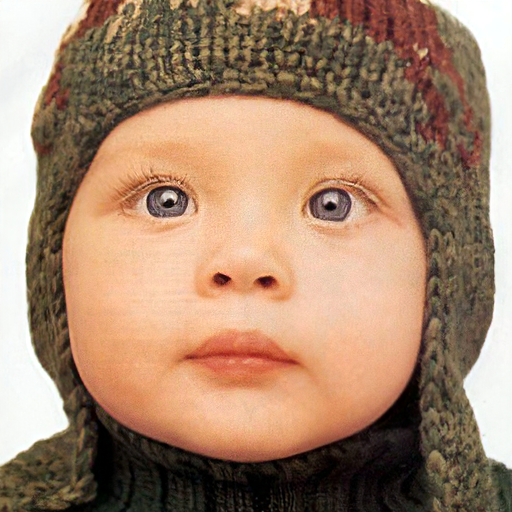} \\
\ESRGAN{} & \MGAN{} & \tarsier{}  \\
\end{tabular}
\caption{\small 
\tarsier{} compared to baselines on the eye of the boy in set5.
Compared to \MGAN{} and to other baselines, the image generated by \tarsier{} is sharper.}
\vspace{1ex}
\label{fig:boy_crop}
\end{center}
\end{minipage}
\begin{minipage}{0.1\textwidth}

\end{minipage}
\begin{minipage}{0.45\textwidth}
\begin{center}
\includegraphics[width=0.9\textwidth]{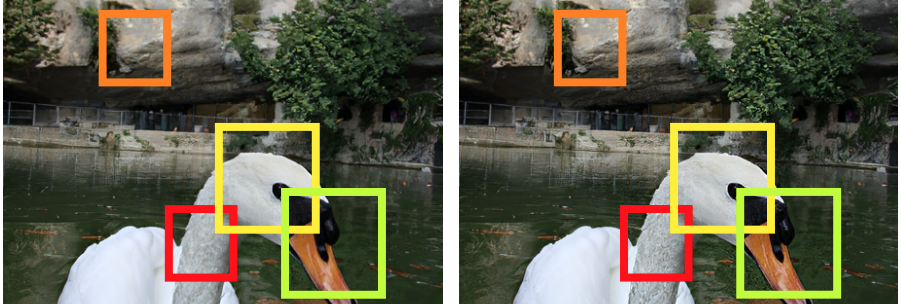}\\
\includegraphics[width=0.9\textwidth]{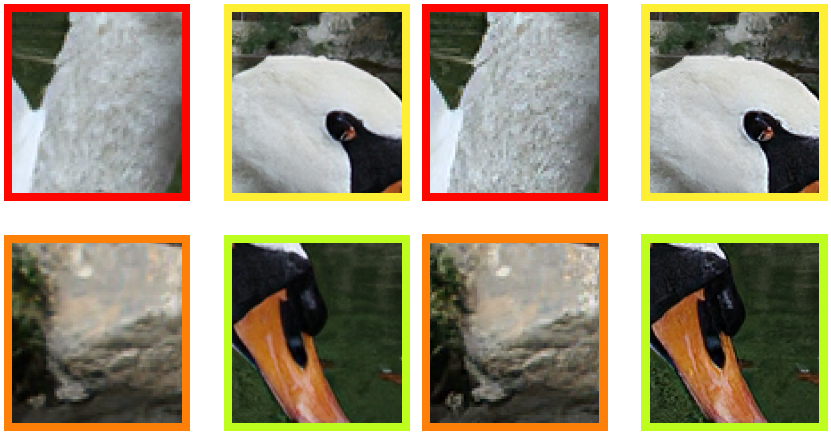}\\
\MGAN{} ~~~~~~~~~~~~~~ \tarsier{} ~~~
\omitme{
  \begin{minipage}{0.4\columnwidth}
    \textbf{\MGAN{}}\par\medskip
      \begin{tikzpicture}[zoomboxarray, zoomboxes below, zoomboxarray inner gap=0.2cm, zoomboxarray columns=2, zoomboxarray rows=2]
    \node [image node] { \includegraphics[trim={0 100 0 100}, clip, width=\textwidth]{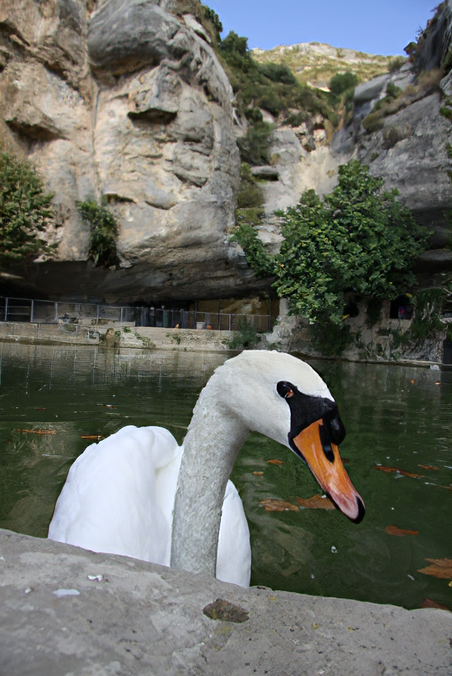} };
    \zoombox[color code=red]{0.45,0.28}
    \zoombox[magnification=2,color code=yellow]{0.6,0.42}
    \zoombox[color code=orange]{0.3,0.7}
    \zoombox[magnification=2,color code=lime]{0.75,0.28}
\end{tikzpicture}
\end{minipage}
  \begin{minipage}{0.4\columnwidth}
    \textbf{Tarsier}\par\medskip
      \begin{tikzpicture}[zoomboxarray, zoomboxes below, zoomboxarray inner gap=0.2cm, zoomboxarray columns=2, zoomboxarray rows=2]
    \node [image node] { \includegraphics[trim={0 100 0 100}, clip, width=\textwidth]{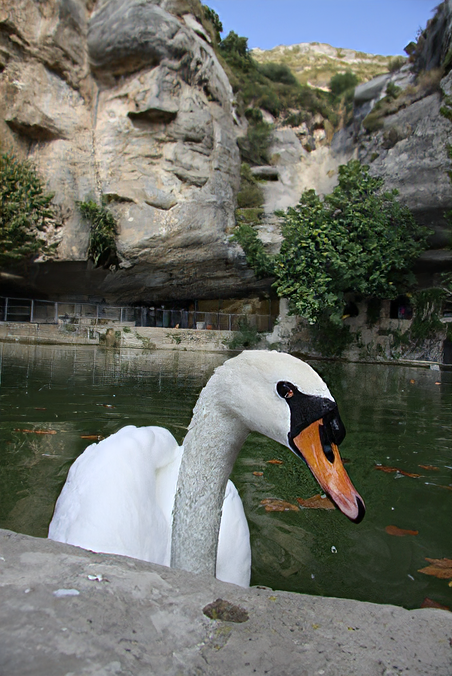} };
    \zoombox[color code=red]{0.45,0.28}
    \zoombox[magnification=2,color code=yellow]{0.6,0.42}
    \zoombox[color code=orange]{0.3,0.7}
    \zoombox[magnification=2,color code=lime]{0.75,0.28}
\end{tikzpicture}
\end{minipage}}
\end{center}
\caption{\small Left: baseline with noise set to zero at inference time (\MGAN{}). Right: Our result, where some details on the image optimized with \dcma{} are sharper than on the baseline image, notably the neck, eye, and beak.
}
\label{fig:swan}
\end{minipage}
\end{figure}

\section{Introduction}
Super-resolution has received much attention from the computer vision and machine learning communities and enjoys a wide range of applications in domains such as medical imaging ~\cite{greenspan2009super, oktay2017anatomically}, security~\cite{rasti2016convolutionalsecurity, nguyen2018super} and other computer vision tasks  ~\cite{dai2016isimage,haris2018task,shermeyer2019effects,noh2019better_srpreprocessing}.
Several architectures were proposed to maximize the Peak Signal-to-Noise Ratio (PSNR) ~\cite{dong2015image, kim2016accurate, lai2017deep, SANdai2019second}. 
However, the PSNR score contradicts quality assessments from human observers and PSNR-oriented methods tend to produce blurry images\bricpr{~\cite{blau2018PIRM,rad2019srobb}}. 
Recent works\bricpr{~\cite{wang2018esrgan,malagan, zhang2019ranksrgan}} evaluate their models based on the PIRM \br{perceptual index}, which  \bricpr{combines the MA~\cite{ma2017learning} and NIQE~\cite{mittal2012making} scores and} is related to perceptual quality.

Methods based on Generative Adversarial Networks (GANs) are especially successful at producing sharp and realistic images. Among them, we can list Super Resolution GANs (SRGANs)~\cite{ledig2017photo}, and follow up works ~\cite{malagan,wang2018esrgan,park2018srfeat, sajjadi2017enhancenet, ledig2017photo} that \br{ perform well according to the PIRM criterion}. 
In this paper, we improve SRGAN and its variant known as \MGAN{}~\cite{malagan} that uses noise injection~\cite{noiseinjection} at training time.
We consider the noise injection as a free parameter that can be leveraged in order to improve the output quality. 
More precisely, at inference time, we optimize the noise injection using an evolutionary algorithm in order to produce sharper and more detailed images.
Our novel objective function combines an Image Quality Assessment tool, \koncept\cc{~\cite{koncept512reference}}, and the discriminator of the GAN in order to improve both image quality and realism.
Thanks to our \ccci{proposed}\omitme{pessimistic} score, our criterion is also robust across datasets and requires little hyper-parameter tuning. 

When optimizing our objective function, gradient-based optimization methods classically used in deep learning (e.g. Adam, Gradient Descent) quickly get stuck on critical points and only result in marginal improvements in our criterion.
Evolutionary methods~\cite{holland} are known as the \otcyo{jeep of artificial intelligence~\cite{rob1,rob2}}: they are compatible with rugged objective functions without gradient and search for flat, stable optima~\cite{notinoculation}. 
Moreover, \omitme{other works}\ccci{the work} \cite{camilleinspir} optimizing the latent space of GANs \ccci{finds} that evolutionary methods are especially robust to imperfect surrogate objective functions. 
Our experiments also support the use of evolutionary methods for optimizing noise injection in GANs: they show that \dcma{} is well suited to optimizing our rugged objective criterion, as it outperforms gradient-based methods on many datasets.
When optimizing our criterion with an evolution strategy, we produce sharper and more detailed images. We outperform \MGAN{} and other Super Resolution methods quantitatively (i.e. according the PIRM perceptual index) and qualitatively \brfour{(according to a human study)}. \bricpr{A short paper~\cite{roziere2020evolutionary} presented the idea of merging super-resolution GAN and quality estimation. The present paper contains experimental results, equations and detailed algorithms.}

\section{Background}

\subsection{\MGAN{}: noise injection in super-resolution GANs}

SRGAN~\cite{ledig2017photo}, an application of conditional GANs to super-resolution, showed that GANs are well suited to improving the perceptual quality of images generated with Super Resolution.
SRGAN \ccci{is} equipped with attention in~\cite{asrgan}, and made size-invariant thanks to global pooling in~\cite{gpsrgan}. Sometimes extended with dense connections, it performs well for perception-related measures~\cite{dsrgan}.
\ESRGAN ~\cite{wang2018esrgan} is an enhancement based on a novel architecture containing blocks without batch normalization layers~\cite{ioffe2015batch}, use of Relativistic average GAN~\cite{jolicoeur2018relativistic} and features before activation for the perceptual loss. \MGAN{}~\cite{malagan} is an extension of \ESRGAN{} using additional residual connections and noise injection \cc{as depicted in Fig.~\ref{fig:figure_tarsier}}. Gaussian noise is added to the output of \otcicpr{each of the 23 residual layers of each of the 3 blocks (total 69 layers)} along with learned per-feature scaling factors. 
It leverages stochastic variations that randomize only local aspects of the generated images without changing our global perception of these images\ccci{, in a spirit similar to}~\cite{karras2018style}.
The \cc{injected noise} $z$ is usually set to \ccc{zero} at inference time. In the present paper, we consider $z\neq 0$, optimized by evolutionary algorithms based on an objective function built with \koncept{} and the discriminator.

\subsection{\koncept: image quality metric via supervised deep learning}

We modify our conditional GAN to improve the quality of the generated images by considering the \koncept~\cite{koncept512reference} image quality assessment (IQA) model.

\omitme{\koncept{} was built as follows:
\begin{itemize}
    \item A dataset, KonIQ-10k, was sourced and subjectively annotated by crowdsourcing, with technical quality ratings.
    \item A deep-learning model was trained on this dataset.
\end{itemize}}
\ccc{\koncept{} is built by training a deep-learning model to rate the quality of images from the KonIQ-10k dataset.}
\ccc{The }KonIQ-10k \ccc{dataset} is the largest, reliably annotated in-the-wild IQA publicly available dataset, consisting of 10,073 subjectively rated images, \mtb{each  rated} by 120 users. 
The images are sourced from Flickr and selected to cover a wide range of content sources (object categories) and quality-related indicators. 
The domain of the images in KonIQ-10k is particularly suitable for evaluating super-resolution methods. 
\br{As the images on Flickr are predominantly taken by} amateur photographers, they often focus incorrectly, take hand-held shots producing motion-blurs, \br{or take pictures} in low-light conditions. 

For brevity, the \koncept{} IQA model will be referred to as \ccc{$\K512$. I}t takes as input an image $I$ and outputs its estimated quality score $K(I)$. \koncept{} is highly accurate, with a performance on the KonIQ-10k test set equivalent to the mean opinion score coming from nine users. However, when \ccc{$\K512$} is cross-tested on another in-the-wild IQA database (LIVE in-the-wild), implying some domain shift, the performance drops; in this case, the model is equivalent to roughly the opinion of a single user. 
Thus, \ccc{$\K512$} \brtwo{is effective on} in-the-wild images and may not fare well on synthetic degradation. The limitation\brtwo{s} of the model require careful consideration when deploying it for guided super-resolution, as done in the present work.

\section{\tarsier: optimized noise injection}

\begin{figure}[t]
    \begin{minipage}{0.45\textwidth}
    \begin{center}
    \textbf{Training Stage (\MGAN{})}
    \medskip
    \includegraphics[width=\columnwidth]{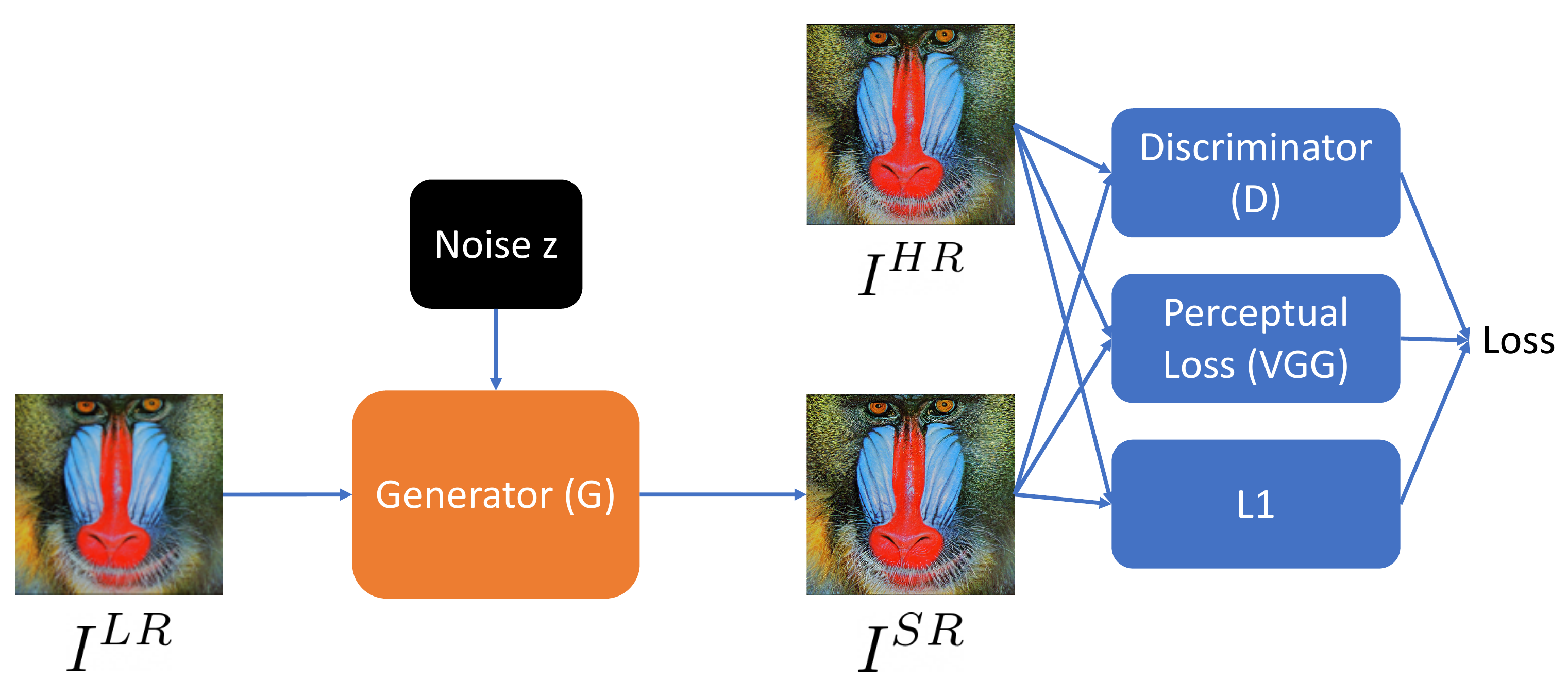}
    \end{center}
    \end{minipage}
    \begin{minipage}{0.45\textwidth}
    \centering
    \textbf{Inference Stage (\tarsier{})}
    \medskip
    \includegraphics[width=\columnwidth]{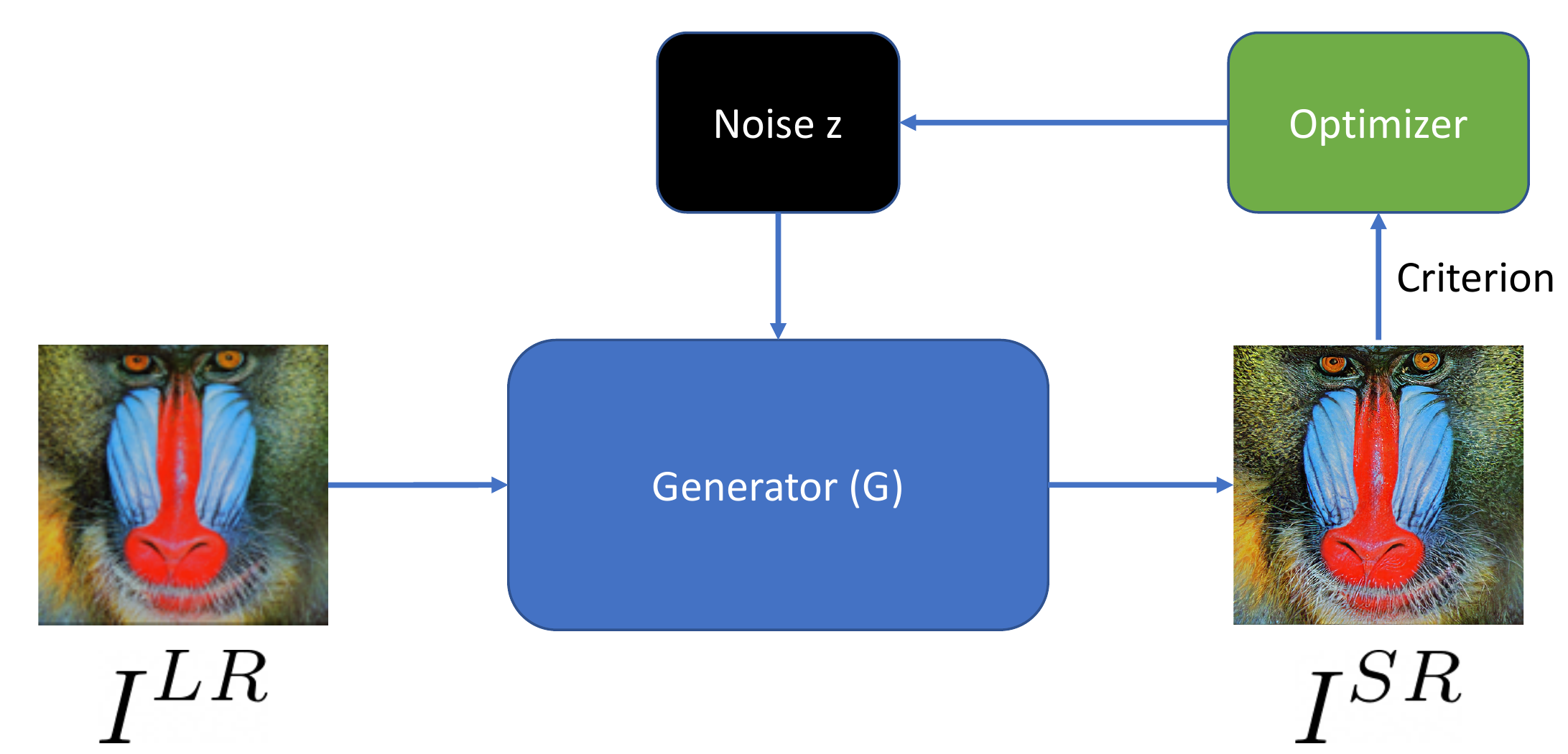}
    \end{minipage}
    \caption{\small \ccci{Top}: \MGAN{} training. A Gaussian noise $z$ is injected at training time. At test time, $z$ is set to zero. \ccci{Bottom}: our \tarsier{} noise optimization method. We use a pre-trained \MGAN{} generator to generate high resolution images from low resolution images and a noise chosen using an optimizer \ccci{that maximizes our criterion}}
    \label{fig:figure_tarsier}
\end{figure}

Given a low-resolution image $I^{LR}$ and a noise vector $z$ \br{of dimension $d$,} \br{we can use a trained \MGAN{} generator $G$ to generate a high resolution image $I^{SR} = G(z, I^{LR})$. }
Whereas GANs with noise injection \br{typically use} $z=0$, our algorithm \tarsier{} (named after a family of haplorrhine primates with excellent eyesight) considers $z=z^*$ maximizing 
one of the criteria defined below. \ccc{See Figure~\ref{fig:figure_tarsier} for an overview of our approach.}

\subsection{Image Quality Score}
\label{sec:Image_quality_score}
A naive quality scorer could consist in simply applying \koncept{} to the output of \MGAN:
$$S_q^{Naive} = \K512(G(z, I^{LR})).$$
We observe that the scores given by $\K512$ remain accurate as long as the generated images remain similar to natural images or to the images generated by \MGAN.
However, the images obtained by optimizing for $S_q^{Naive}$ are not in this category\br{: they} tend to contain many artifacts, which increase\cc{s} the score of $\K512$ despite being unrealistic and visually displeasing.
It prompted us to optimize a \emph{pessimistic} version of the score given by $\K512$:
\begin{equation}
  S_q(z) = L^+(\K512(G(z, I^{LR}))-\K512(G(0, I^{LR})))
  \label{eq:sq}
\end{equation}
with 
 \begin{equation}
    L^+(x)=
    \begin{cases}
    \log(1+x)&\mbox{if }x>0\\
    x & \mbox{otherwise.}
    \end{cases}
    \label{eq:lplus}
\end{equation}

We call $S_q$ pessimistic because it trusts bad scores \br{while taking} a \mtb{safety margin} for good ones.
As $\K512(G(0, I^{LR}))$ is constant, $S_q$ is equivalent to $S_q^{Naive}$ when the proposed $z$ fares worse than the zero-noise injection baseline (i.e. $\K512(G(z, I^{LR}))-\K512(G(0, I^{LR})) < 0$). In other words, we consider that $S_q^{Naive}$ is an accurate evaluation of the quality of the generated image when it gives a \brtwo{poor} score.
Inversely, when the proposed $z$ fares better than the baseline (i.e. $\K512(G(z, I^{LR}))-\K512(G(0, I^{LR})) > 0$), $S_q$ increases logarithmically with the score of $\brtwo{\K512}$ in order to account for the possibility \br{that the image quality may be overestimated}. Moreover, $S_q$ increases similarly to $S_q^{Naive}$ for $\K512(G(z, I^{LR}))$ very close to the baseline score, while it increases slowly when it is much larger.

Another advantage of the logarithm in the $L^+$ function is that it makes our optimization process less scale-dependent when the proposed $z$ clearly outperforms the baseline: when \brtwo{$x = \K512(G(z, I^{LR}))-\K512(G(0, I^{LR}))$} is large,  multiplying $x$ by a constant is almost equivalent to adding a constant to our score and does not change the behaviour of our optimizer. This property makes our hyper-parameters more robust across images and datasets.

\subsection{Realism Score}
Optimizing for $S_q$ compels the generator to produce high-quality image\brtwo{s}, but nothing forces it to generate realistic images.
Generative Adversarial Networks train a generator concurrently with a discriminator, which aims at evaluating the realism of a high-resolution image given an input low-resolution image. The later can be used to produce a realism score.
The discriminator still suffers from the same issue as \cc{$\K512{}$}. It \ccci{is} trained on real high-resolution images and images generated by the \MGAN{} generator\brtwo{. Since} the noise is sampled from a Gaussian distribution centered on zero during training\brtwo{,} the discriminator is inaccurate on images generated with noises having extreme norms or directions.
In order to mitigate this issue, we \mtb{define} a pessimistic realism score similarly to the quality score defined in Section \ref{sec:Image_quality_score}: 
\begin{equation}
  S_r(z) = L^+(\discr(G(z, I^{LR}))-\discr(G(0, I^{LR}))),
  \label{eq:sreal}
\end{equation}
where $\discr(I)$ is the discriminator score for the image $I$.

In practice, the discriminator can only take images of dimension $128\times128$ as input and the high-resolution images we generate with \MGAN{} are always of higher dimension. To compute $S_r$ on the whole image, we divide the image in patches of size $128\times128$ and compute the score on each one of them. In order to ensure that each part of the image looks realistic, we define $S_r$ as the \mtb{minimum score} over all the patches.

\subsection{Final Criterion}
We \brtwo{define} our final criterion, which is to be maximized, by adding a $l_2$ penalization to the quality score (\otc{E}q. \ref{eq:sq}) and the realism score (\otc{E}q. \ref{eq:sreal}) with suitable coefficients:
\begin{equation}
    \mathcal{C}_1(z) = \lambda_q S_q(z) + \lambda_r S_r(z)-\frac{\lambda_p}{d}||z||_2^2 ,
    \label{eq1}
\end{equation}
where $d$ is the dimension of $z$, and $\lambda_q, \lambda_r, \lambda_p$ are scale factors for each term.
\ccc{The application of this first criterion} yields better results on blurry images. \ccc{This observation motivates the definition of a second criterion that adapts the noise injection to the blurriness of the input.} \brtwo{We define it as follows:}
\begin{equation}
    \mathcal{C}_2(z) = \lambda_q S_q(z) + \lambda_r S_r(z)-\frac{\lambda_p B(G(0, I^{LR}))}{d}||z||_2^2 ,
    \label{eq2}
\end{equation}
where $B(I)$ is the standard deviation of the Laplacian of image $I$, divided by $\sqrt{1000}$ in order to keep the regularization on the same order of magnitude. The value of $B(I)$ increases when the blurriness of image $I$ decreases. \ccc{Maximizing }Eq.~\ref{eq2} increases the regularization when the images are less blurry. 


We \bricpr{tested} several values of the hyperparameters $\lambda_q$, $\lambda_r$ and $\lambda_p$. With our pessimistic scores, \tarsier{} \ccc{is} not very sensitive to small variations of the hyperparameters. While some values \ccc{lead} to slightly better results on some of the datasets, we \ccc{find} the results obtained by setting $\lambda_q=\lambda_r=\lambda_p=1$ to be satisfactory and particularly robust across datasets.

\ccci{
\tarsier{} uses Diagonal Covariance Matrix Adaptation (DCMA)~\cite{diagcma} for optimizing our criteria. CMA Evolution Strategy~\cite{hansen2001completely, hansen2006cma} is a second order method which estimates a covariance matrix closely related to the Hessian matrix in Quasi-Newton methods.
It requires no tedious parameter tuning, since the choice of most of the internal parameters of the strategy is automatically done during training with methods such as Cumulative Step-Size Adaptation~\cite{arnold2004performance}. DCMA is a variant of CMA in which the covariance matrix is constrained to be diagonal. It reduces the time and space complexity of CMA from quadratic to linear. It evaluates fewer parameters for the covariance matrix and requires fewer function evaluations than CMA for estimating the covariance in high dimension. }

\section{Experiments}

\subsection{Setup}
\br{As in~\cite{malagan}, we use an \MGAN{} model trained on DIV2K~\cite{agustsson2017ntire}, \bricpr{improving significantly the performance of ESRGAN by injecting noise at training time. Noise vectors are injected after each residual connection in every residual dense block, and are sampled randomly from a Gaussian distribution at training time. At inference time, \tarsier{} learns noise vectors for each residual connection.}} 
\otcicpr{We \bricpr{consider them as} additional degrees of freedom for \bricpr{further improving the performance}, as measured by \bricpr{our criterion.}} The dimension of the noise is $d=$27,600.
\br{We use the code and the weights of the \koncept{} model available online\footnote{\url{https://github.com/subpic/koniq}}. We compute the \tarsier{} criterion on a Tesla V100 GPU.}
\brtwo{We run $\times4$ upscaling experiments on widely used super-resolution datasets: Set5\cite{bevilacqua2012low}, Set14\cite{zeyde2010single}, the PIRM Validation and Test datasets\cite{blau2018PIRM}, Urban100\cite{huang2015single}, and OST300\cite{wang2018recovering}}.
\br{We compute the PIRM score on Matlab using the code made available for the PIRM2018 challenge\footnote{\url{https://github.com/roimehrez/PIRM2018}}.}
We refer to~\cite{wang2018esrgan, ma2017learning,mittal2012making} and references therein \otc{ for \br{a} precise definition of the perceptual quality estimation used in the present document.}
Here is a description of the optimizers we assess to maximize Eq. \ref{eq1} and \ref{eq2}.

\subsection{Optimizer choice: evolutionary computation}
We use evolutionary algorithms implemented in the nevergrad library~\cite{nevergrad}. The motivations for this choice are:
\begin{itemize}[leftmargin=*]
\setlength\itemsep{0em}
        \item As pointed out in~\cite{notinoculation} and \cite{camilleinspir} in the context of computer vision,  evolutionary algorithms provide solutions robust to imperfect objective functions. More precisely, by focusing on optima stable by random variable-wise perturbations, evolutionary algorithms behave well on the real objective function (in particular, human assessment) when we optimize a proxy (here, our criterion). \otc{We see }\ccc{in Table~}\otc{\ref{table:loss_comparison2} that, even from a purely numerical point of view, evolutionary computation outperforms gradient-based algorithms in the present setting. We observe that gradient-based methods tend to get stuck in suboptimal critical points after a few \mtb{hundreds  iterations}. Gradient descent slightly outperforms Adam on all datasets but its performances are still way below these of \dcma{}.}

\begin{table}[t]
\caption{\small\label{table:loss_comparison2} Comparison of derivative-free methods (CMA, \dcma{}, the 1+1 evolution strategy \cc{\cite{schumer1968adaptive}}, Differential Evolution) and gradient-based methods (Adam and Gradient Descent) for optimizing $\mathcal{C}_1$ (higher is better). Gradient-based methods tend to get stuck in sub-optimal local minima and do not perform as well as \dcma{}. } 
\begin{center}
\begin{adjustbox}{max width=\columnwidth}
\begin{tabular}{llll}
\toprule
Dataset/Method&Set5 &Set14 & PIRM Val\\
\midrule
Random Search & $1.28\pm 0.65\%$& $0.61 \pm 0.50$ & $0.06\pm 0.03$ \\
CMA & $1.75 \pm 1.36$ & $0.71 \pm 0.68$ & $0.28 \pm 0.16$\\
DCMA & $\textbf{4.51}\pm 0.86$ & $\textbf{3.80} \pm 0.50$ & $\textbf{3.13}\pm 0.13$\\
(1+1) & $3.74 \pm 2.1$ & $1.63 \pm 1.36$ & $0.01 \pm 0.02$\\
GD & $1.79 \pm 0.77$ & $0.85 \pm 0.54$ & $0.93\pm0.18$\\
ADAM & $1.51 \pm 0.77$ & $0.84 \pm 0.63$ & $0.82\pm 0.23$\\
\bottomrule
\end{tabular}
\end{adjustbox}
\end{center}
\end{table}

\begin{table}[t]
\setlength{\tabcolsep}{4pt}
\begin{center}
\caption{\small\label{table:comparison_PIRM_DFO_Grad} Gradient-based methods also do not perform as well according to the PIRM perceptual score ~\cite{blau2018PIRM} (lower is better). The budget is 10000 function evaluations. For each optimizer, the Blur version optimizes $\mathcal{C}_2$ (\otc{E}q. \ref{eq2}) and the other version optimizes $\mathcal{C}_1$ (Eq. \ref{eq1}). \omitme{For each line, the b}Best result shown in bold, second best underlined.}
\begin{tabular}{lllllll}
\toprule
& DCMA & DCMA  & GD & GD  & ADAM   & ADAM  \\
&    & + Blur &       &  + Blur  &        &    +Blur           \\
\midrule
set5& \underline{2.787} & \textbf{2.667} & 3.033 & 3.026& 2.997 & 2.998\\
set14 & \underline{2.740}& \textbf{2.656} &2.826& 2.828& 2.937   & 2.937        \\
PIRM Val& \underline{2.348} & \textbf{2.335}& 2.376& 2.375  & 2.399   & 2.399 \\
PIRM Test & \textbf{2.260} & \underline{2.277} &2.297 & 2.297 &2.307   & 2.307 \\
\bottomrule
\end{tabular}
\end{center}
\end{table}

\begin{table}[t]
\setlength{\tabcolsep}{3pt}
    \begin{center}
        \caption{\small PIRM perceptual scores computed (lower is better) on the raw outputs of several super-resolution models. For each dataset, the best result is shown in bold and the second best is underlined. \tarsier{} optimizes $\mathcal{C}_1$ (Eq. \ref{eq1}) while \tarsier{} + Blur optimizes $\mathcal{C}_2$ (Eq. \ref{eq2}). Both \tarsier{} and \tarsier+Blur outperform \MGAN{} in all cases.}
    \begin{adjustbox}{max width=\columnwidth}
    \begin{tabular}{l c c c c c c}
    \toprule
        Dataset & SAN\cite{SANdai2019second} &\EnhanceNet & \ESRGAN& \MGAN{} & \tarsier & \tarsier\\
                & & & & & & + Blur\\
    \midrule
        set5& 5.94&2.93 & 3.76 & 3.21 & \underline{2.79} & \textbf{2.67}\\
        set14& 5.37 &3.02 & 2.93 & 2.80 & \underline{2.74} & \textbf{2.66}\\
        PIRM Val&-&2.68& 2.55& 2.37 & \underline{2.35}& \textbf{2.34}\\
        PIRM Test& - &2.72 & 2.44 & 2.29 & \textbf{2.26}& \underline{2.28}\\
        Urban100 & 5.12&\textbf{3.47} &3.77 & 3.55 &3.50 & \underline{3.49}\\
        OST300 &-&2.82 &2.49 & 2.49 & \textbf{2.47}& \underline{2.47}\\
    \bottomrule
    \end{tabular}
    \end{adjustbox}
    \label{tab:results_PIRM}
    \end{center}
\end{table}

\begin{figure}
    \centering
    \begin{tabular}{ccc}
    \includegraphics[width=0.45\columnwidth]{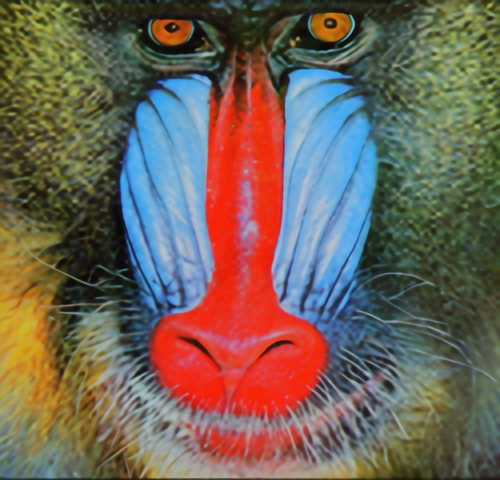}&
     \includegraphics[width=0.45\columnwidth]{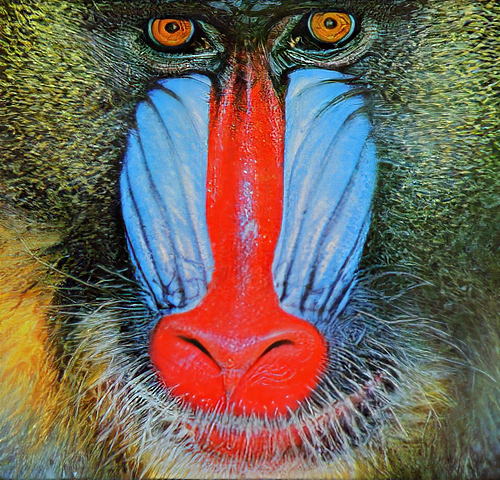}\\
     SAN & \tarsier{}
    \end{tabular}{}
    \caption{\small\bricpr{SAN and \tarsier{} on the baboon image from Set 14. The image generated by SAN is visibly more blurry than that generated by \tarsier{}. It is also the case for other images and when comparing SAN to \ESRGAN{}.}}
    \label{fig:comparison_SAN_ESRGAN_Tarsier_baboon}
\end{figure}

    \item There is no need for gradient estimation. \br{Although our criteria are differentiable, evolutionary methods could also be used to optimize non-differentiable criteria (e.g., direct human feedback). Moreover, computing the overall gradient over distinct deep learning frameworks can be a burden, and evolutionary methods do not require it.}
    \omitme{We implement gradient-based methods for the sake of comparison in \br{this} paper\br{. They do not perform as well as \dcma{}}}. 
    \item The optimization is naturally parallel.
\end{itemize}
We mainly use \dcma{}~\cite{diagcma} as a derivative-free optimization algorithm because it is fast and reliable. Compared to CMA, the diagonal covariance matrix reduces the computational cost \otc{ and reduces the budget requirement as we do not have to evaluate an entire covariance matrix. Compared to the $(1+1)$-evolution strategy with one-fifth rule, \br{DCMA has} anisotropic mutations\br{: it adapts the step-size on each dimension}.}
\otcicpr{
	DCMA frequently \bricpr{ranks high} in black-box optimization benchmarks, \bricpr{particularly when} the problem is partially separable and/or when the dimension is large. This is definitely the case here with 27600 parameters, \bricpr{while the black-box optimization literature focuses on }
	dimension\bricpr{s} $<200$\cite{bbob}. DCMA and Differential Evolution (DE) \ccci{are} often \bricpr{the most performant} among algorithms not using specific decompositions, e.g. on Nevergrad's dashboard\cite{dashboard} or on LSGO\cite{lsgo}. In our experiments, DCMA performed better than DE.
	}


\subsection{Results}

Our \cc{main quantitative} experimental results are presented in Table~\ref{table:comparison_PIRM_DFO_Grad} and Table~\ref{tab:results_PIRM}. In Table \ref{table:comparison_PIRM_DFO_Grad}, we compare gradient-based methods to the evolutionary method that performed best according to our criterion: \dcma{}. \dcma{} outperforms both GD and ADAM on four standard super-resolution datasets and when optimizing either the Blur or the normal criterion. 
Using the results from Tables~\ref{table:loss_comparison2} and \ref{table:comparison_PIRM_DFO_Grad}, we decided to eliminate gradient-based methods and to use \dcma{} to optimize our criterion in \tarsier{}.

\paragraph{Perceptual score}
\br{In Table \ref{tab:results_PIRM},} we use the PIRM score to compare our method, \tarsier, to \MGAN{}\cite{malagan}, \ESRGAN{}\cite{wang2018esrgan}, \EnhanceNet{}\cite{sajjadi2017enhancenet} \cc{and SAN \cite{SANdai2019second}}.
All our baselines are \br{perception-driven} approaches based on GANs\cc{, except SAN that is based on a convolution network with attention.}
\ESRGAN{} outperforms \EnhanceNet{} on every dataset except set5, which is much smaller than the others, and Urban100, on which \EnhanceNet{} is particularly \cc{efficient}. \MGAN{} outperforms \ESRGAN{} on every dataset, even though it is only by a small margin for OST300. However, \EnhanceNet{} still surpasses \MGAN{} on set5 and Urban100.
Whether it uses the Blur or the non-Blur criterion, \tarsier{} outperforms \MGAN{} on every dataset. It also outperforms every other method on every dataset except Urban100, on which \EnhanceNet{} is slightly better. 
\bricpr{SAN performs well in terms of PSNR, but not in terms of perceptual scores. The images it generates appear much \ccci{blurrier} than those generated by \tarsier{} or any of our baselines, see Fig.~\ref{fig:comparison_SAN_ESRGAN_Tarsier_baboon}}.

\begin{figure}
\setlength{\tabcolsep}{2pt}
\begin{minipage}{0.45\textwidth}
    \centering
    \includegraphics[width=.49\columnwidth]{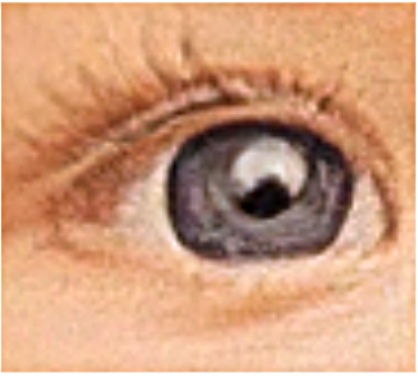}
    \includegraphics[width=.49\columnwidth]{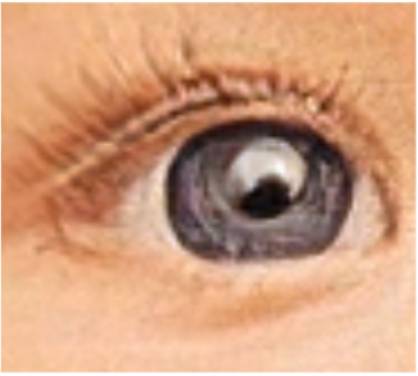}
    
    \caption{\small On the left, unsharp masking 
    applied to the output of \MGAN. On the right, the output of \tarsier. We set the \textit{amount}, \textit{radius}, and \textit{threshold} parameters of unsharp masking to 80\%, 5 pixels and 0 respectively.}
    \label{eye}
\end{minipage}\hfill
\begin{minipage}{0.45\textwidth}
    \centering
    \begin{tabular}{p{15mm}cc}
         &$\lambda_r=0$& $\lambda_r = 1$\\
         Raw \newline Score & \includegraphics[align=c, trim={75 100 25 0},clip,width=.3\columnwidth]{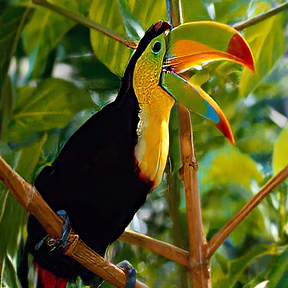}
         & \includegraphics[align=c, trim={75 100 25 0},clip,width=.3\columnwidth]{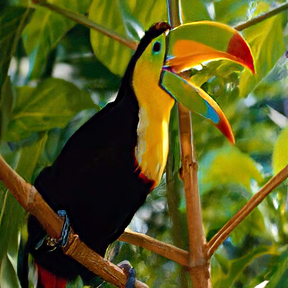} \\
         Pessimistic \newline Score& 
         \includegraphics[align=c,trim={75 100 25 0},clip,width=.3\columnwidth]{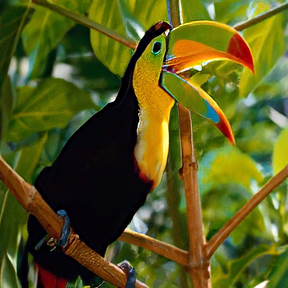} 
         & \includegraphics[align=c,trim={75 100 25 0},clip,width=0.3\columnwidth]{{images/set5/bird_koncept_512_DiagonalCMA_blur_None}.png}
    \end{tabular}

    \caption{\small Ablation study for the pessimistic score and the realism loss. Tarsier is on the bottom right. } 
    \label{fig:bird_pessimistic}
    \hfill
    \end{minipage}
\end{figure}

\paragraph{Qualitative comparison}
We present \br{some} examples in Figs.~\ref{fig:boy_crop} and~\ref{fig:swan}. 
 \br{Compared to \MGAN{}, \tarsier{} generates images that are sharper and more detailed. For instance, Tarsier produces sharper and more natural eyelashes on the boy in Fig. \ref{fig:boy_crop}. 
On Fig. \ref{fig:swan}, it is capable of generating sharper patterns and shadows on the stones behind the swan, as well as more convincing wet feathers on the neck of the animal. The beak also appears less blurry.}
We show in Fig. \ref{eye} that \tarsier{} does more than applying a classical sharpening filer (unsharp masking) on top of \MGAN{}, while keeping the image clean, and reducing graininess. 

\brfour{\paragraph{Human study} We conducted a double-blind human study to validate that \tarsier{} improves the images generated by \MGAN{}. We took random samples of 20 images for the PIRM Test dataset and 30 for the OST300 dataset. We generated high resolution images with \MGAN{} and \tarsier{}, and asked human reviewers to pick the best image among the two images shown in a random order. As \MGAN{} already generates high-quality images, it is difficult to compare its output to that of \tarsier{} without zooming on the image. In order to make the comparison easier for the annotators, we generated zoomed-in images automatically for most different sub-images (measured using the PSNR). The images generated using \tarsier{} \ccci{are} preferred in $75.0\%$ of the cases for PIRM Val and $76.7\%$ for OST300. On both sets, \tarsier{} significantly outperforms \MGAN{} (p-value $<$ 0.05).}

\subsection{Ablation Study}

\begin{figure}[t]
\centering
\begin{tabular}{cc}
\includegraphics[width=.23\textwidth]{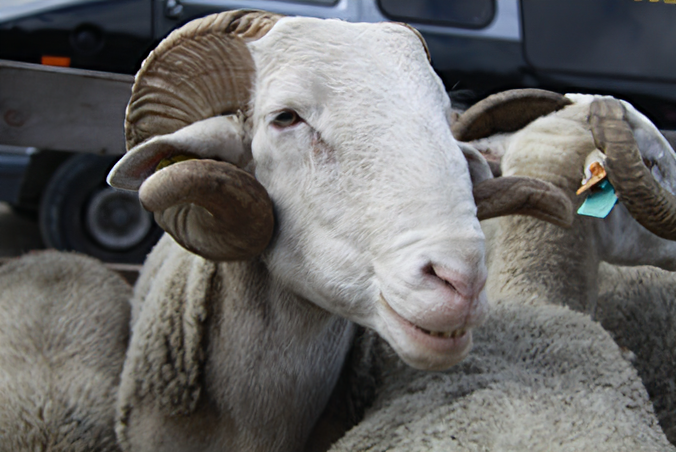}& \includegraphics[width=0.23\textwidth]{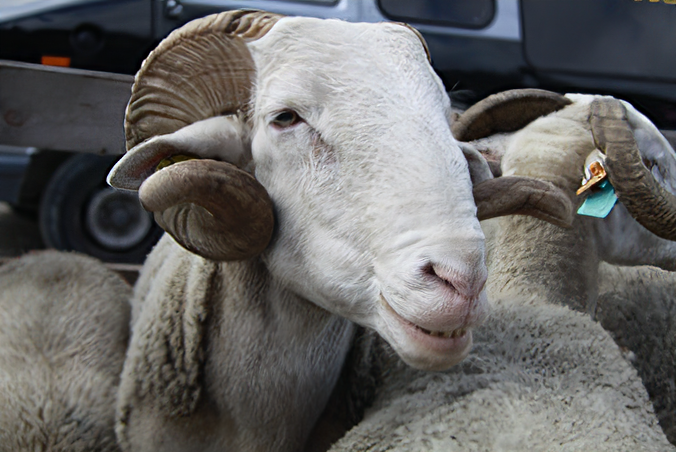}\\
\MGAN{} & Our result with \tarsier{} 
\end{tabular}\\
~ \\
\begin{tabular}{ccc}
\includegraphics[trim={150 200 310 50},clip,width=.13\textwidth]{{images/PIRM_VAL/32_koncept_512_Baseline}.png}&
\includegraphics[trim={150 200 310 50},clip,width=.13\textwidth]{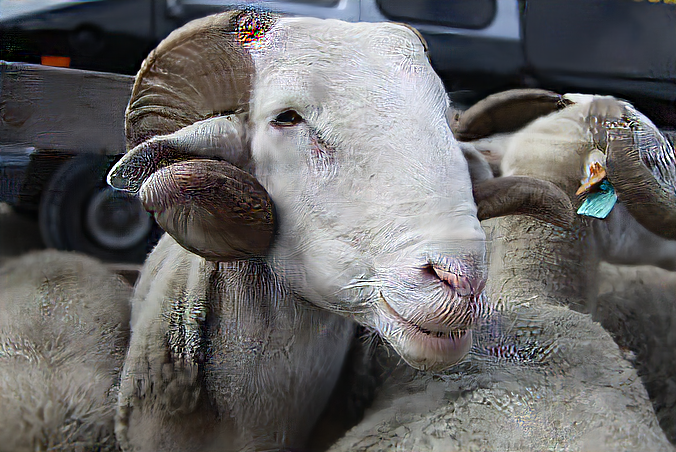}&
\includegraphics[trim={150 200 310 50},clip,width=.13\textwidth]{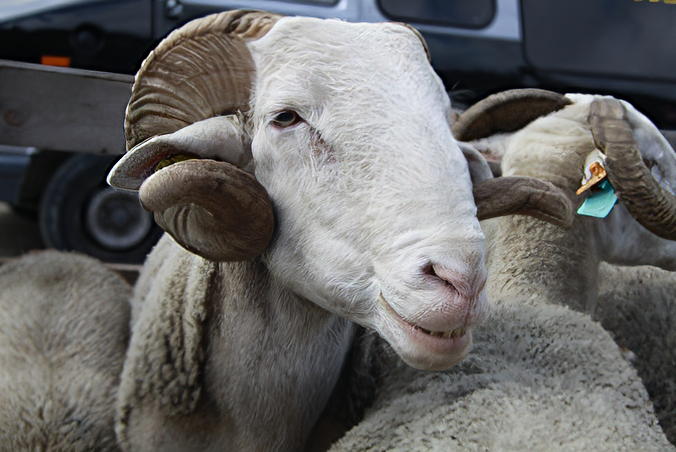}
\\
\MGAN{} & $\lambda_r = \lambda_p =0$ & $\lambda_r =0$ \\
\includegraphics[trim={150 200 310 50},clip,width=.13\textwidth]{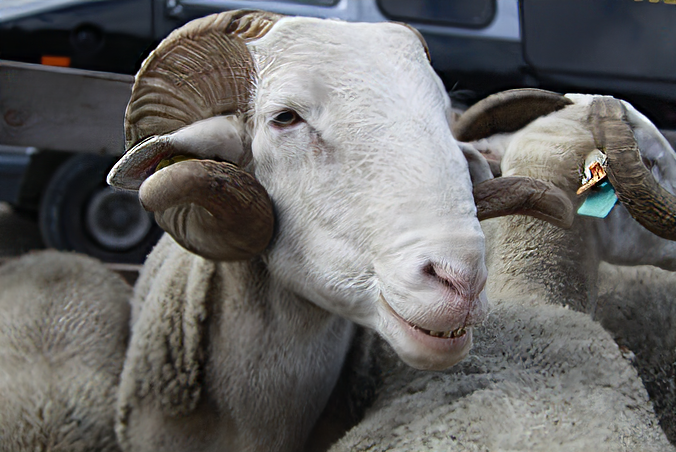}&
\includegraphics[trim={150 200 310 50},clip,width=.13\textwidth]{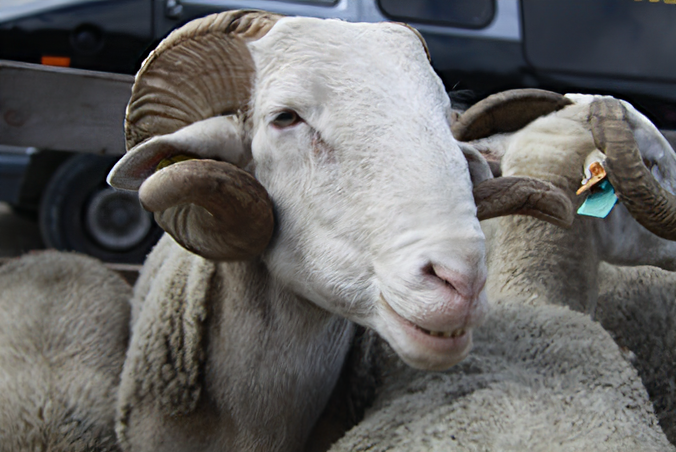}&
\includegraphics[trim={150 200 310 50},clip,width=.13\textwidth]{{images/PIRM_VAL/32_koncept_512_DiagonalCMA_blur_None}.png} \\
$\lambda_p =0$ & $\lambda_q =0$ & \tarsier  \\
\end{tabular}
        \caption{\small Ablation study on an image from the PIRM validation dataset. }
\label{fig:ablation_sheep}

\end{figure}

\begin{figure}[htb]
\centering
\begin{tabular}{cc}
\includegraphics[width=.23\textwidth]{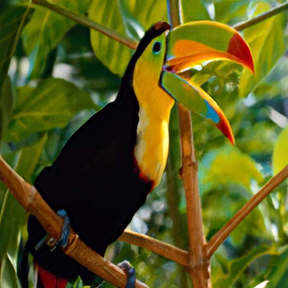}& \includegraphics[width=0.23\textwidth]{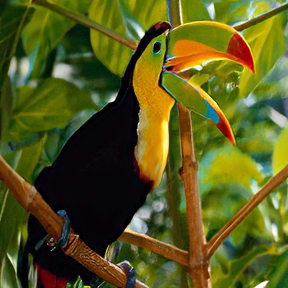}\\
\MGAN{} & Our result with \tarsier
\end{tabular}\\
~ \\
\begin{tabular}{cccc}
\includegraphics[trim={75 100 25 0},clip,width=.30\columnwidth]{{images/set5/bird_koncept_512_Baseline}.png}&
\includegraphics[trim={75 100 25 0},clip,width=.30\columnwidth]{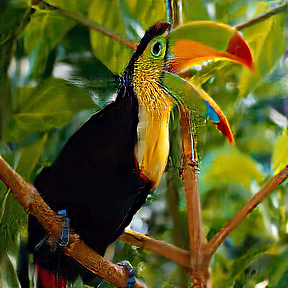}&
\includegraphics[trim={75 100 25 0},clip,width=.30\columnwidth]{{images/set5/ablation/bird_koncept_512_DiagonalCMA_penal1.0_27600_disc0.0_clampingNone_bud10000_blur_None}.png}\\
\MGAN{} & $\lambda_r = \lambda_p =0$ & $\lambda_r =0$ \\
\includegraphics[trim={75 100 25 0},clip,width=.30\columnwidth]{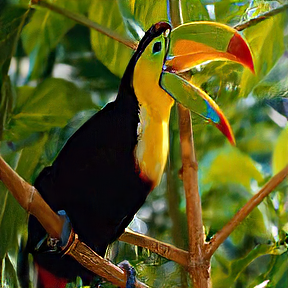}&
\includegraphics[trim={75 100 25 0},clip,width=.30\columnwidth]{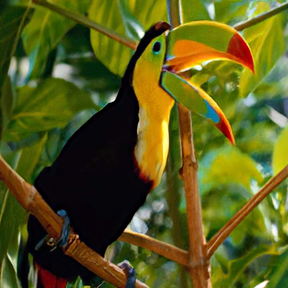}&
\includegraphics[trim={75 100 25 0},clip,width=.30\columnwidth]{{images/set5/bird_koncept_512_DiagonalCMA_blur_None}.png} \\
$\lambda_p =0$ & $\lambda_q =0$ & \tarsier  \\
\end{tabular}
        \caption{\small Ablation study on an image from the set5 dataset. Removing the realism score and the penalization produces heavy artifacts. Removing only the penalization term still produces visible artifacts. Removing quality score makes the image blurrier and less detailed.}
\label{fig:ablation_bird}

\end{figure}

We study \br{the impact of each term in our criterion by setting the corresponding parameters to 0. We observe that removing both the realism score and the penalization ($\lambda_r = \lambda_p =0$) leads to the generation of images with unrealistic and visually displeasing artifacts (see Fig.~\ref{fig:ablation_sheep}). Both terms act as regularizers: if one of \brtwo{them} is set to 1, \brtwo{most of the artifacts disappear.}} 
\brthree{Removing only the penalization term ($\lambda_p =0$) sometimes produces slightly more detailed images (e.g., sheep on Fig.~\ref{fig:ablation_sheep}). However, it can also produce some artifacts (see bird on Fig. \ref{fig:ablation_bird}). We decided to keep the penalization term for more robustness.
\brtwo{Removing the realism score ($\lambda_r =0$) frequently produces small artifacts (see Fig.~\ref{fig:ablation_bird}).}
Without the quality term ($\lambda_q=0$), the images become more blurry and similar to those obtained with $z=0$ (\MGAN{}).}


\br{As expected, the logarithm in the definition of the quality (Eq.~\ref{eq:sq}) and realism (Eq.~\ref{eq:sreal}) scores can act as a regularizer and keep the generated images realistic. It is particularly visible when we also remove the realism score from our criterion (see $\lambda_r =0$ on Fig. \ref{fig:bird_pessimistic}). The artifacts can also be avoided by increasing the penalization coefficient. \brtwo{However, without the pessimistic loss, the right value for $\lambda_r$ \ccci{would} depend on the dataset and even on the image within a dataset. Therefore, using the raw instead of the pessimistic scores \ccci{would} make the method less robust and the penalization factor difficult and costly to tune.} The pessimistic loss is less \mtb{scale-}dependent \brtwo{and} the results it produces are much more robust across images and datasets: \brtwo{all the results we present \ccci{are} obtained with all the parameters set to 1.}}
\br{Another benefit of the pessimistic loss is to allow us to keep the quality and realism scores on the same scale. With the raw scores, i.e. without applying a logarithm to positive relative scores, the optimizer tends to optimize for whichever criterion is easiest to optimize. In our case, the optimizer} often \ccc{reaches} very high realism scores (median value of $99.83$ on set5 and set14) and almost \ccc{ignores} the quality score (median value of $4.45$). 
\br{Despite that, we} \ccc{do} \br{not often observe artifacts on the images generated without the pessimistic score. It indicates that optimizing the realism score does not easily create artifacts. However, these images are \ccci{blurrier} since the quality term \ccci{tends to be ignored} (see right column on Fig. \ref{fig:bird_pessimistic}). \brfour{With the pessimistic scores, the median value of the quality score becomes $2.42$ on set5 and set14, meaning the median raw scores goes up from $4.45$ to $11.25$.}}

\section{Conclusion}

Noise injection \ccc{has been left unexploited by super-resolution approach\brtwo{es} that} typically use noise $z=0$ at inference time. \ccc{In this work,} \br{we combine \ccc{the perceptual image quality \brtwo{assessment model} \koncept{} and the adversarial network output} \omitme{discriminator of the GAN and a penalization }into a novel criterion accounting for the image's technical quality and realism. 
Without retraining the model, we optimize $z$ using our criterion and outperform \ccci{the state-of-the-art} on several standard super-resolution datasets \brthree{according to the PIRM score and to human opinion.}}

\br{Our experiments show that, in addition to being easy to use, \dcma{} is better suited to our problem than gradient-based methods. On every dataset we tested, it outperform\brtwo{ed} gradient-based methods for optimizing the criterion and when compared using the PIRM score.}


Though this optimized noise injection \ccci{is} applied to super-resolution in the present paper, the method can be applied for optimizing noise injection in general.
\br{In this paper, we \ccci{choose} \koncept{} for our quality score in order to \mtb{optimize the} technical quality of the images. We could use the same method \brtwo{on} another type of criterion (e.g. the artistic quality).}

\bibliographystyle{abbrv}
\bibliography{bibliodoe,bibliolsca,biblio}

\begin{thebibliography}{10}

\bibitem{agustsson2017ntire}
E.~Agustsson and R.~Timofte.
\newblock Ntire 2017 challenge on single image super-resolution: Dataset and
  study.
\newblock In {\em Computer Vision and Pattern Recognition Workshops}, 2017.

\bibitem{arnold2004performance}
D.~V. Arnold and H.-G. Beyer.
\newblock Performance analysis of evolutionary optimization with cumulative
  step length adaptation.
\newblock {\em IEEE Transactions on Automatic Control}, 2004.

\bibitem{bevilacqua2012low}
M.~Bevilacqua, A.~Roumy, C.~Guillemot, and M.~L. Alberi-Morel.
\newblock Low-complexity single-image super-resolution based on nonnegative
  neighbor embedding.
\newblock 2012.

\bibitem{blau2018PIRM}
Y.~Blau, R.~Mechrez, R.~Timofte, T.~Michaeli, and L.~Zelnik-Manor.
\newblock The 2018 pirm challenge on perceptual image super-resolution.
\newblock In {\em Proceedings of the European Conference on Computer Vision
  (ECCV)}, 2018.

\bibitem{dai2016isimage}
D.~Dai, Y.~Wang, Y.~Chen, and L.~Van~Gool.
\newblock Is image super-resolution helpful for other vision tasks?
\newblock In {\em Winter Conference on Applications of Computer Vision (WACV)}.
  IEEE, 2016.

\bibitem{SANdai2019second}
T.~Dai, J.~Cai, Y.~Zhang, S.-T. Xia, and L.~Zhang.
\newblock Second-order attention network for single image super-resolution.
\newblock In {\em Conference on Computer Vision and Pattern Recognition}, 2019.

\bibitem{dong2015image}
C.~Dong, C.~C. Loy, K.~He, and X.~Tang.
\newblock Image super-resolution using deep convolutional networks.
\newblock {\em transactions on pattern analysis and machine intelligence},
  2015.

\bibitem{gpsrgan}
S.~Gautam, D.~K. Pradhan, P.~C. Chhipa, and S.~Nakajima.
\newblock Microgan: Size-invariant learning of gan for super-resolution of
  microscopic images.
\newblock 2019.

\bibitem{greenspan2009super}
H.~Greenspan.
\newblock Super-resolution in medical imaging.
\newblock {\em The computer journal}, 2009.

\bibitem{hansen2006cma}
N.~Hansen.
\newblock The {CMA} evolution strategy: a comparing review.
\newblock In {\em Towards a new evolutionary computation}. 2006.

\bibitem{bbob}
N.~Hansen, A.~Auger, S.~Finck, and R.~Ros.
\newblock Real-parameter black-box optimization benchmarking: Experimental
  setup.
\newblock Technical report, Universit{\'{e}} Paris Sud, INRIA Futurs,
  {\'{E}}quipe TAO, Orsay, France, Mar.~24, 2012.

\bibitem{hansen2001completely}
N.~Hansen and A.~Ostermeier.
\newblock Completely derandomized self-adaptation in evolution strategies.
\newblock {\em Evolutionary computation}, 2001.

\bibitem{haris2018task}
M.~Haris, G.~Shakhnarovich, and N.~Ukita.
\newblock Task-driven super resolution: Object detection in low-resolution
  images.
\newblock {\em arXiv preprint 1803.11316}, 2018.

\bibitem{holland}
J.~H. Holland.
\newblock Genetic algorithms and the optimal allocation of trials.
\newblock {\em {SIAM} J. Comput.}, 2(2):88--105, 1973.

\bibitem{koncept512reference}
V.~Hosu, H.~Lin, T.~Sziranyi, and D.~Saupe.
\newblock Koniq-10k: An ecologically valid database for deep learning of blind
  image quality assessment.
\newblock {\em IEEE Transactions on Image Processing}, 2020.

\bibitem{huang2015single}
J.-B. Huang, A.~Singh, and N.~Ahuja.
\newblock Single image super-resolution from transformed self-exemplars.
\newblock In {\em Conference on Computer Vision and Pattern Recognition}, 2015.

\bibitem{ioffe2015batch}
S.~Ioffe and C.~Szegedy.
\newblock Batch normalization: Accelerating deep network training by reducing
  internal covariate shift.
\newblock {\em International Conference on Machine Learning}, 2015.

\bibitem{jolicoeur2018relativistic}
A.~Jolicoeur-Martineau.
\newblock The relativistic discriminator: a key element missing from standard
  {GAN}.
\newblock {\em ICLR}, 2019.

\bibitem{notinoculation}
K.~A.~D. Jong.
\newblock Genetic algorithms are not function optimizers.
\newblock In {\em Foundations of Genetic Algorithms}, volume~2 of {\em
  Foundations of Genetic Algorithms}, pages 5 -- 17. Elsevier, 1993.

\bibitem{karras2018style}
T.~Karras, S.~Laine, and T.~Aila.
\newblock A style-based generator architecture for generative adversarial
  networks.
\newblock In {\em Conference on Computer Vision and Pattern Recognition}, 2019.

\bibitem{kim2016accurate}
J.~Kim, J.~Kwon~Lee, and K.~Mu~Lee.
\newblock Accurate image super-resolution using very deep convolutional
  networks.
\newblock In {\em Conference on Computer Vision and Pattern Recognition}, 2016.

\bibitem{lai2017deep}
W.-S. Lai, J.-B. Huang, N.~Ahuja, and M.-H. Yang.
\newblock Deep laplacian pyramid networks for fast and accurate
  super-resolution.
\newblock In {\em Conference on Computer Vision and Pattern Recognition}, 2017.

\bibitem{ledig2017photo}
C.~Ledig, L.~Theis, F.~Husz{\'a}r, J.~Caballero, A.~Cunningham, A.~Acosta,
  A.~Aitken, A.~Tejani, J.~Totz, Z.~Wang, et~al.
\newblock Photo-realistic single image super-resolution using a generative
  adversarial network.
\newblock In {\em Conference on Computer Vision and Pattern Recognition}, 2017.

\bibitem{ma2017learning}
C.~Ma, C.-Y. Yang, X.~Yang, and M.-H. Yang.
\newblock Learning a no-reference quality metric for single-image
  super-resolution.
\newblock {\em Computer Vision and Image Understanding}, 2017.

\bibitem{rob2}
N.~Milano, P.~Pagliuca, and S.~Nolfi.
\newblock Robustness, evolvability and phenotypic complexity: Insights from
  evolving digital circuits.
\newblock {\em Evolutionary Intelligence}, pages 83--95, 2019.

\bibitem{mittal2012making}
A.~Mittal, R.~Soundararajan, and A.~C. Bovik.
\newblock Making a “completely blind” image quality analyzer.
\newblock {\em Signal Processing Letters}, 2012.

\bibitem{nguyen2018super}
K.~Nguyen, C.~Fookes, S.~Sridharan, M.~Tistarelli, and M.~Nixon.
\newblock Super-resolution for biometrics: A comprehensive survey.
\newblock {\em Pattern Recognition}, 2018.

\bibitem{noh2019better_srpreprocessing}
J.~Noh, W.~Bae, W.~Lee, J.~Seo, and G.~Kim.
\newblock Better to follow, follow to be better: Towards precise supervision of
  feature super-resolution for small object detection.
\newblock In {\em Proceedings of the IEEE International Conference on Computer
  Vision}, 2019.

\bibitem{oktay2017anatomically}
O.~Oktay, E.~Ferrante, K.~Kamnitsas, M.~Heinrich, W.~Bai, J.~Caballero, S.~A.
  Cook, A.~De~Marvao, T.~Dawes, D.~P. O‘Regan, et~al.
\newblock Anatomically constrained neural networks (acnns): application to
  cardiac image enhancement and segmentation.
\newblock {\em IEEE transactions on medical imaging}, 2017.

\bibitem{park2018srfeat}
S.-J. Park, H.~Son, S.~Cho, K.-S. Hong, and S.~Lee.
\newblock Srfeat: Single image super-resolution with feature discrimination.
\newblock In {\em European Conference on Computer Vision}, 2018.

\bibitem{asrgan}
H.~N. Pathak, X.~Li, S.~Minaee, and B.~Cowan.
\newblock Efficient super resolution for large-scale images using attentional
  {GAN}.
\newblock In {\em 2018 International Conference on Big Data}, 2018.

\bibitem{rad2019srobb}
M.~S. Rad, B.~Bozorgtabar, U.-V. Marti, M.~Basler, H.~K. Ekenel, and J.-P.
  Thiran.
\newblock Srobb: Targeted perceptual loss for single image super-resolution.
\newblock In {\em Proceedings of the IEEE International Conference on Computer
  Vision}, pages 2710--2719, 2019.

\bibitem{malagan}
N.~C. {Rakotonirina} and A.~{Rasoanaivo}.
\newblock Esrgan+ : Further improving enhanced super-resolution generative
  adversarial network.
\newblock In {\em International Conference on Acoustics, Speech and Signal
  Processing (ICASSP)}, pages 3637--3641, May 2020.

\bibitem{nevergrad}
J.~Rapin and O.~Teytaud.
\newblock {Nevergrad - A gradient-free optimization platform}.
\newblock \url{https://GitHub.com/FacebookResearch/Nevergrad}, 2018.

\bibitem{dashboard}
J.~Rapin and O.~Teytaud.
\newblock {Nevergrad's Dashboard}.
\newblock \url{https://GitHub.com/FacebookResearch/Nevergrad}, 2018.

\bibitem{rasti2016convolutionalsecurity}
P.~Rasti, T.~Uiboupin, S.~Escalera, and G.~Anbarjafari.
\newblock Convolutional neural network super resolution for face recognition in
  surveillance monitoring.
\newblock In {\em International conference on articulated motion and deformable
  objects}. Springer, 2016.

\bibitem{camilleinspir}
M.~Riviere, O.~Teytaud, J.~Rapin, Y.~LeCun, and C.~Couprie.
\newblock Inspirational adversarial image generation.
\newblock {\em arXiv preprint 1906.11661}, 2019.

\bibitem{diagcma}
R.~Ros and N.~Hansen.
\newblock A simple modification in cma-es achieving linear time and space
  complexity.
\newblock In G.~Rudolph, T.~Jansen, N.~Beume, S.~Lucas, and C.~Poloni, editors,
  {\em Parallel Problem Solving from Nature -- PPSN X}, pages 296--305, Berlin,
  Heidelberg, 2008. Springer Berlin Heidelberg.

\bibitem{roziere2020evolutionary}
B.~Roziere, N.~C. Rakotonirina, V.~Hosu, H.~Lin, A.~Rasoanaivo, O.~Teytaud, and
  C.~Couprie.
\newblock Evolutionary super-resolution.
\newblock In {\em Proceedings of the 2020 Genetic and Evolutionary Computation
  Conference Companion}, pages 151--152, 2020.

\bibitem{sajjadi2017enhancenet}
M.~S. Sajjadi, B.~Scholkopf, and M.~Hirsch.
\newblock Enhancenet: Single image super-resolution through automated texture
  synthesis.
\newblock In {\em International Conference on Computer Vision}, 2017.

\bibitem{rob1}
J.~{Schonfeld} and D.~A. {Ashlock}.
\newblock A comparison of the robustness of evolutionary computation and random
  walks.
\newblock In {\em Proceedings of the 2004 Congress on Evolutionary Computation
  (IEEE Cat. No.04TH8753)}, volume~1, pages 250--257 Vol.1, 2004.

\bibitem{schumer1968adaptive}
M.~Schumer and K.~Steiglitz.
\newblock Adaptive step size random search.
\newblock {\em IEEE Transactions on Automatic Control}, 1968.

\bibitem{shermeyer2019effects}
J.~Shermeyer and A.~Van~Etten.
\newblock The effects of super-resolution on object detection performance in
  satellite imagery.
\newblock In {\em Conference on Computer Vision and Pattern Recognition
  Workshops}, 2019.

\bibitem{noiseinjection}
C.~K. Sønderby, J.~Caballero, L.~Theis, W.~Shi, and F.~Huszár.
\newblock Amortised map inference for image super-resolution, 2016.

\bibitem{wang2018recovering}
X.~Wang, K.~Yu, C.~Dong, and C.~Change~Loy.
\newblock Recovering realistic texture in image super-resolution by deep
  spatial feature transform.
\newblock In {\em Conference on Computer Vision and Pattern Recognition}, 2018.

\bibitem{wang2018esrgan}
X.~Wang, K.~Yu, S.~Wu, J.~Gu, Y.~Liu, C.~Dong, Y.~Qiao, and C.~C. Loy.
\newblock Esrgan: Enhanced super-resolution generative adversarial networks.
\newblock In {\em European Conference on Computer Vision Workshops (ECCVW)},
  2018.

\bibitem{dsrgan}
Z.~Wang, K.~Jiang, P.~Yi, Z.~Han, and Z.~He.
\newblock Ultra-dense gan for satellite imagery super-resolution.
\newblock {\em Neurocomputing}, 2019.

\bibitem{lsgo}
A.~Zamuda, J.~Brest, B.~Bošković, and V.~Zumer.
\newblock Large scale global optimization using differential evolution with
  self-adaptation and cooperative co-evolution.
\newblock pages 3718 -- 3725, 07 2008.

\bibitem{zeyde2010single}
R.~Zeyde, M.~Elad, and M.~Protter.
\newblock On single image scale-up using sparse-representations.
\newblock In {\em International conference on curves and surfaces}. Springer,
  2010.

\bibitem{zhang2019ranksrgan}
W.~Zhang, Y.~Liu, C.~Dong, and Y.~Qiao.
\newblock Ranksrgan: Generative adversarial networks with ranker for image
  super-resolution.
\newblock In {\em Proceedings of the IEEE International Conference on Computer
  Vision}, pages 3096--3105, 2019.

\end{thebibliography}

\end{document}